# AquaSAM: Underwater Image Foreground Segmentation

Muduo Xu[1,2], Jianhao Su[3], and Yutao Liu[1]

[1] Faculty of Information Science and Engineering, Ocean University of China
[2] Heriot-Watt University, Edinburgh, UK mx2003@hw.ac.uk
[3] University of Dundee, UK 2441423@dundee.ac.uk

**Abstract.** The Segment Anything Model (SAM) has revolutionized natural image segmentation, nevertheless, its performance on underwater images is still restricted. This work presents AquaSAM, the first attempt to extend the success of SAM on underwater images with the purpose of creating a versatile method for the segmentation of various underwater targets. To achieve this, we begin by classifying and extracting various labels automatically in SUIM dataset. Subsequently, we develop a straightforward fine-tuning method to adapt SAM to general foreground underwater image segmentation. Through extensive experiments involving eight segmentation tasks like human divers, we demonstrate that AquaSAM outperforms the default SAM model especially at hard tasks like coral reefs. AquaSAM achieves an average Dice Similarity Coefficient (DSC) of 7.13 (%) improvement and an average of 8.27 (%) on mIoU improvement in underwater segmentation tasks.

## 1 Introduction

Semantic segmentation is a notable problem in the domains of computer vision [1][2][3] for its usefulness in estimating scene geometry, inferring interactions and spatial relationships among objects, salient object identification, and more. Underwater image segmentation is a fundamental task in underwater imaging analysis which involves identifying and delineating regions of interest (ROI) in various underwater images. Accurate segmentation is indispensable for many applications, including underwater image detection, classification and underwater image enhancement (UIE).

Recently, there have been significant advancements in the field of natural image segmentation with the support of segmentation foundation models [4][5][6]. These models enable accurate and efficient segmentation of objects in a fully automatic or interactive manner. Typically based on transformer architectures, these models leverage pre-trained weights to achieve state-of-the-art performance and demonstrate an unprecedented ability to generalize across a wide range of natural images.

The first and most notable segmentation foundation model is the Segment Anything Model (SAM) [7]. SAM is trained on over 1 billion masks and possesses strong capabilities for generating accurate object masks based on prompts such as bounding boxes, points, or texts, as well as in a fully automatic manner. However, the applicability of these models to underwater image segmentation remains limited due to significant discrepancies between natural images and target-oriented underwater images. Several studies have demonstrated that SAM may struggle with typical underwater image segmentation tasks [8], [9], [10], [11], [12], [13], [14], as well as other challenging scenarios [15], [16], [17], [18], [33], [34], [35], particularly when the targets have weak boundaries. This outcome is unsurprising since SAM's training set primarily comprises natural and highquality image datasets where objects typically possess strong edge information.

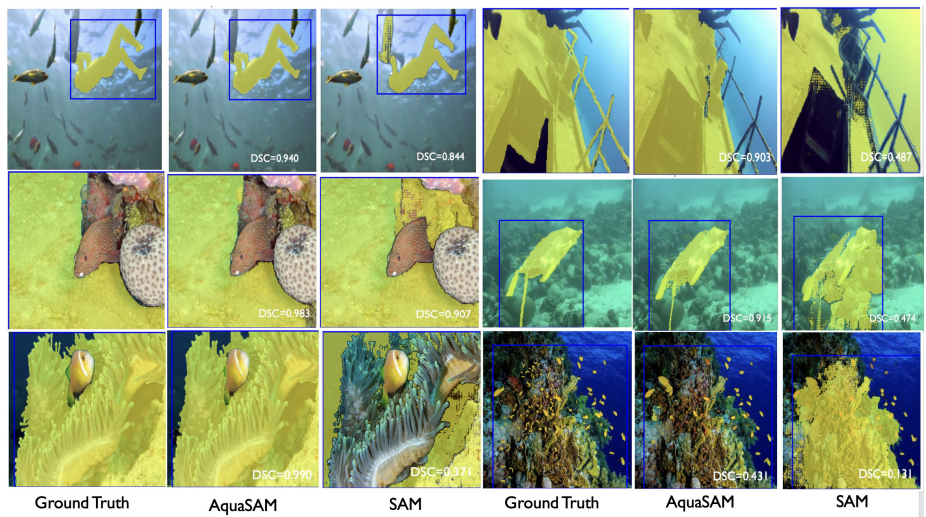

**Fig.1.** Visualized Samples of the AquaSAM model and pre-trained SAM model segmentation results. AquaSAM significantly improves the segmentation performance across various modalities, depth and segmentation tasks.

In this study, we present AquaSAM, the first attempt to adapt the Segment Anything Model (SAM) to the underwater domain for universal image segmentation. Drawing inspiration from SAM's robust capacity, primarily achieved through large-scale supervised training, we begin by classifying and extracting various labels automatically in SUIM dataset. We analyze the network architecture components of SAM and evaluate their potential usefulness in image segmentation tasks. Lastly, we propose a fine-tuning approach to adapt SAM specifically for underwater image segmentation. Our experiments, encompassing 8 image segmentation tasks, demonstrate significant performance improvements achieved by our method in the realm of underwater image segmentation when compared to SAM alone.

## 2 Related Work

### 2.1 Underwater Image Segmentation

Underwater image segmentation has gained significant attention in recent years due to its wide range of applications in underwater robotics, marine biology, and underwater archaeology. Various machine learning methods have been proposed to tackle the challenges associated with underwater imagery, such as low visibility, color distortion, and texture degradation. In this section, we present a comprehensive review of the state-of-the-art (SOTA) techniques and their achievements in underwater image segmentation. Early approaches to underwater image segmentation primarily relied on traditional computer vision techniques, such as thresholding, region-based methods, and clustering algorithms. These methods often struggled to handle the unique characteristics of underwater images, resulting in limited segmentation accuracy. As a result, researchers turned their attention towards machine learning methods, which demonstrated better performance in capturing the complex relationships within underwater image data.

One popular approach in underwater image segmentation is based on deep learning techniques. Deep learning has shown remarkable success in various computer vision tasks and have been adapted to handle the challenges of underwater imagery. For instance, Drews-Jr, P. et al. (2021) proposed the first work to use a CNN approach to underwater image segmentation in the wild given the real underwater images in the wild and their respective ground truths. [19] Similarly, Arain. et al. introduced incoporating feature-based stereo matching with learning-based segmentation to produce a more robust obstacle map [20] which considers direct binary learning of the presence or absence of underwater obstacles.

To address the distorted color issue in underwater images, several works have focused on color correction techniques. Zhang et al. proposed a color balance strategy which balances the color differences between channel a and channel b in the CIELAB color space [21], effectively improving the performance of underwater image segmentation. In a similar vein, Li et al. [22] introduced an underwater color image segmentation method via RGB color channel fusion to obtain the grayscale image with high foreground-background contrast and improves color space accuracy.

Furthermore, researchers have explored the integration of multi-modal information, such as depth maps and polarization images, to enhance and access underwater image segmentation performance. Islam et al. [23] presented a deep residual model which balances the trade-off between performance and computational efficiency, providing competitive performance while ensuring fast inference.

## 2.2    Segment Anything Model (SAM)

The Segment Anything Model (SAM) incorporates a transformer-based architecture [24], known for its effectiveness in natural language processing [25] and image recognition tasks [26]. SAM utilizes a vision transformer-based image encoder to extract image features and prompt encoders to incorporate user interactions. It further employs a mask decoder to generate segmentation results and confidence scores based on the image embedding, prompt embedding, and output token.

To handle high-resolution images (i.e., 1024×1024), the vision transformer in the image encoder is pre-trained using masked auto-encoder modeling [27]. The resulting image embedding is then downscaled by a factor of 16 (64×64). SAM supports four different prompts: points, boxes, texts, and masks. Each point is encoded using Fourier positional encoding [28] and two learnable tokens to specify foreground and background. The bounding box is encoded by using the point encoding of its top-left and bottom-right corners. Free-form text is encoded using the pre-trained text-encoder in CLIP [29]. The mask prompt maintains the same spatial resolution as the input image and is encoded using convolutional feature maps.

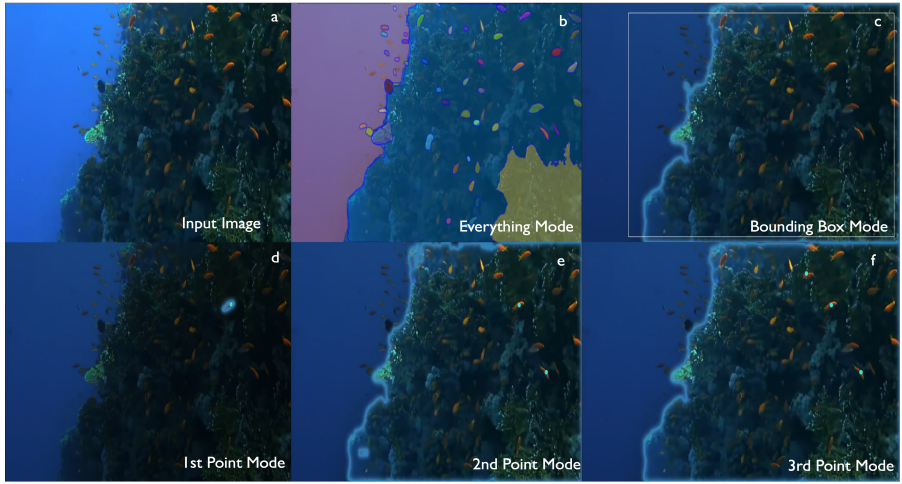

**Fig.2.** Segmentation Results on Underwater Images when applied SAM in different modes

The mask decoder in SAM follows a lightweight design, comprising two transformer layers with a dynamic mask prediction head and an Intersection-overUnion (IoU) score regression head. The mask prediction head generates three 4× downscaled masks, representing the whole object, a part of the object, and a subpart of the object, respectively.

By leveraging the strengths of transformer-based architectures and incorporating various prompt encoders, SAM aims to achieve accurate and versatile segmentation of diverse objects in images. The pretrained image encoder, tailored prompt encoders, and lightweight mask decoder collectively contribute to the effectiveness and efficiency of the Segment-Anything Model.

## 3   Method

### 3.1   Mode Selection in SAM

Three main segmentation modes including everything mode, bounding box mode and point mode are provided by SAM. Fig. 2 shows the three segmentation results of underwater images with different modes. Several regions including fish and grass are divided in the segment-anything mode. However, these segmentation results have limited utility for two primary reasons. Firstly, the obtained result (Fig. 2b) lacks the accuracy of segmentation, a number of fish inside it cannot be segmented, which restricts its interpretability and practical application. Secondly, granted that Segment Anything Model promises that even if it is not trained on underwater datasets, it can perform well in underwater domains, the lack of depth in underwater environments can cause problems. In underwater scenarios, researchers are primarily interested in identifying and analyzing specific regions of interest (ROIs) that hold optical significance, such as the human divers, ruins and coral reefs. Compared with the segment-anything mode, the bounding box mode (Fig. 2c) can perform better in the segmentation of the task-oriented background by just giving the upper-left and bottom-right points. What's more, the point segmentation mode tries to segment the deep fish one by one in the foreground. Although the first-point result (Fig. 2d) looks satisfying, the second-point and the third-point result (Fig. 2e) (Fig.2f) is not guaranteed.

In a nutshell, when employing SAM for underwater image segmentation, the "segment-everything" mode tends to generate partitions that are not useful and ceases to reach high accuracy of segmentation, while the "point-based" mode introduces ambiguity and necessitates multiple iterations for prediction-correction. In contrast, the "bounding box-based" mode enables clear specification of the ROI and produces reasonable segmentation results without the need for repeated trial and error even if the target where hidden in depth can sometimes be unsegmented. Hence, we contend that the bounding box-based segmentation mode holds greater practical value in underwater image segmentation tasks with SAM, compared to the segment-everything and point-based modes.

### 3.2   AquaSAM: Specialized Foundation Models for Underwater Image Segmentation

To adapt SAM for underwater image segmentation, the selection of a suitable user prompt and network component for fine-tuning is crucial. After careful analysis, it is determined that the bounding box prompt is an appropriate choice for precisely specifying the segmentation target. SAM's network architecture comprises three primary components: image encoder, prompt encoder, and mask decoder. Different combinations of these components can be fine-tuned based on specific requirements. Notably, the image encoder, which is built on a vision transformer, incurs the highest computational overhead within SAM. In order to minimize computation costs, we opt to keep the image encoder frozen and focus on fine-tuning the remaining components.

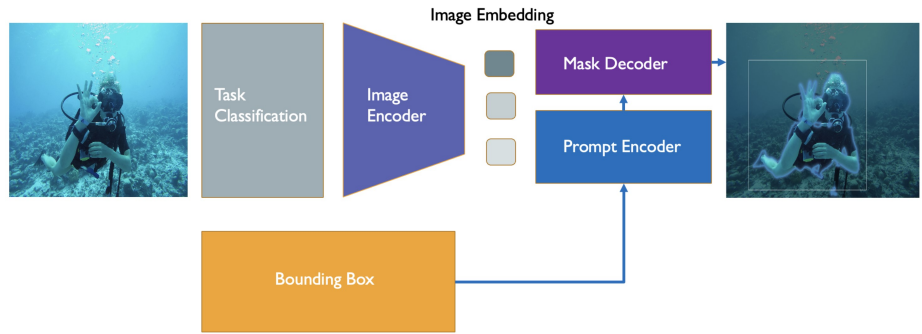

**Fig.3.** AquaSAM: fine-tuning SAM for underwater image segmentation

The images and the corresponding ground truths are classified first and trained seperately which can help guarantee the efficiency of each foreground segmentation task. Given the limited number of underwater images, we begin by curating various tasks to reach higher accuracy. The positional information of the bounding box is encoded by the prompt encoder, which can be reused from the pre-trained bounding-box encoder in SAM. Consequently, we also freeze this component during fine-tuning. Fig. 3 illustrates that the remaining component requiring fine-tuning is the mask decoder. To enhance training efficiency, we can compute the image embedding for all training images beforehand, as the image encoder can be applied prior to prompting the model. This eliminates the need for repetitive computation of the image embedding per prompt. By precomputing the image embedding, significant improvements in training efficiency can be achieved. Furthermore, due to the clarity provided by the bounding box prompt in most scenarios, the mask decoder only needs to generate a single mask rather than three masks when focusing on the foreground specifically in the underwater domains. This is because the bounding box prompt effectively specifies the intended segmentation target.

## 4   Experiments and Results

### 4.1   Data preprocessing

We select SUIM dataset with 8 segmentation tasks to commit further experiments. It includes Background, Human divers, Aquatic plants and sea-grass, Wrecks and ruins, Robots, Reefs and invertebrates, Fish and vertebrates and Sea-floor and rocks. We believe that these foreground and background segmentation task are indispensable for later researches. We handle the preprocessing of underwater images for a specific label (anatomy) using the provided ground truth, underwater images, image size, and the encoder SAM model. Moreover, we apply resizing, thresholding, and other transformations to the input images and masks. By virtue of the salient capacity of SAM encoding model, it generates image embeddings using the model which will return the preprocessed images, masks, and embeddings. Here we focus on performing data preprocessing pipeline for underwater images with various foreground targets, generating labeled data for a specific anatomy using a specified model type and checkpoint. It demonstrates the steps involved in reading, preprocessing, and saving the data for further analysis or model training.

**4.2    Training protocal**

Each dataset was randomly divided into 80 for training and 20 for testing. Segmentation targets with fewer than 100 pixels were excluded from the analysis. We employed a pre-trained ViT-Base model as the image encoder and computed image embeddings offline by inputting normalized images into the image encoder. The image encoder re-scaled the images to a size of 3×1024×1024. During training, the bounding box prompt was generated from the ground-truth mask with a random perturbation of 0-20 pixels. As this work focusing on providing a fundamental segmentation model for underwater image enhancement and the foreground segmentation is indispensable here, the loss function used was the unweighted sum of the Dice loss and cross-entropy loss, which has demonstrated robustness in various segmentation tasks [30][31]. The network was optimized using the Adam optimizer [32] with an initial learning rate of 1e-5.

**4.3    Results on 8 underwater segmentation tasks**

We utilized the Dice Similarity Coefficient (DSC) and mean Intersection over Union (mIoU) as evaluation metrics to assess the region overlap ratio and boundary consensus between the ground truth and segmentation results, which are commonly employed in segmentation tasks [33]. Various comparisons between the pretrained SAM (ViT-B) model and our AquaSAM on 8 foreground and background underwater image segmentation tasks are presented in Table 1. AquaSAM shows significant improvements across all 8 segmentation tasks, achieving an average improvement of 7.13(%) on DSC and an average improvement of
8.27(%) on mIoU. The results of improvements can be further improved greatly when the corresponding dataset is enormous enough.

   While the pretrained SAM model exhibited good performance in single targetoriented segmentation tasks, such as Robots (RO) (Fig. 4e) and Aquatic plants and Sea-grass (PF) (Fig. 4h), it yielded unsatisfactory results for multiple targetoriented foreground segmentation tasks, such as Wrecks and ruins (WR) (Fig. 4c) and Fish and veterbrates (FV) (Fig. 4g). This can be attributed to the fact that the pre-trained SAM model can not adapt the underwater environments especially deep information is needed in certain images. Moreover, the accuracy of segmentation is far from the expected. In contrast, our AquaSAM model outperforms the pretrained SAM model by a significant margin in both DSC and mIoU scores across almost all tasks. Additional segmentation examples are presented in Fig. 4. The pretrained SAM model is particularly prone to producing over-segmentation results, making it challenging to accurately segment targets with weak boundaries. For instance, the pretrained SAM failed to provide accurate segmentation results for Wrecks and ruins (WR) (Fig. 4c) and Fish and vertebrates (FV) (Fig. 4g), even with a relatively tight bounding box prompt. Moreover, when the content inside the bounding box is heterogeneous, the model may struggle to identify the correct segmentation target. Additionally, when segmenting targets with clear boundaries, SAM may generate outliers when the surrounding objects also exhibit good contrasts, as observed in Fish and Vertebrates (FV) segmentations.

**Table 1.** Segmentation performance comparison of SAM and AquaSAM on SUIM dataset

| Task | Mean Dice Score(%) | | | Mean IoU Score(%) | | |
|---|---|---|---|---|---|---|
| | AquaSAM | SAM | Improve (%) | AquaSAM | SAM | Improve (%) |
| **BW** | 87.83 | 83.04 | 5.80 | 82.34 | 76.50 | 7.61 |
| **HD** | 86.32 | 78.35 | 10.14 | 77.82 | 69.63 | 11.76 |
| **WR** | 87.01 | 70.07 | 24.37 | 78.85 | 59.50 | 32.57 |
| **SR** | 77.88 | 74.04 | 5.16 | 68.08 | 64.52 | 5.53 |
| **RO** | 90.02 | 85.63 | 5.39 | 84.08 | 78.85 | 6.65 |
| **RI** | 82.61 | 77.13 | 7.05 | 75.82 | 68.08 | 11.31 |
| **FV** | 77.88 | 52.41 | 32.19 | 68.15 | 45.58 | 49.79 |
| **PF** | 54.87 | 63.06 | -12.95 | 42.92 | 51.43 | -16.48 |

AquaSAM has significantly enhanced the model's ability to identify challenging segmentation targets. Specifically, AquaSAM has demonstrated two major improvements over the pretrained SAM. Firstly, it has improved the model's capability to accurately identify small objects, even in the presence of multiple segmentation targets within the bounding box prompt. Secondly, the model has shown increased robustness towards depth-required underwater image segmentation.

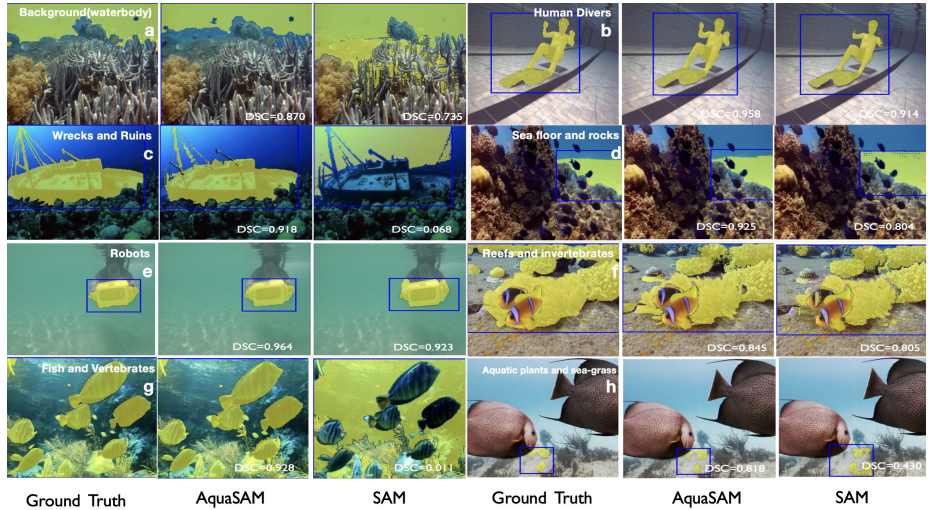

**Fig.4.** More examples of AquaSAM and SAM on various underwater image segmentation tasks

### 4.4   Discussion

We have demonstrated that specifying the oriented tasks and fine-tuning the mask encoder can result in substantial enhancements across different foreground segmentation tasks and image modalities. Nonetheless, the performance of our approach still falls short compared to specialized models designed specifically for underwater image segmentation. Additionally, there is significant potential for improvement in Aquatic plants. Although the results of certain segmentaion tasks may not appear satisfying as the ground truth, Fig. 5. shows that AquaSAM performs better than SAM when applied to multiple segmentation. Moreover, if multiple similar instances surround the segmentation target, a large bounding box can lead to incorrect segmentation results.

To address the limitations of AquaSAM, we believe that leveraging larger models and increasing the dataset size would be beneficial given that now there is no such a large dataset for underwater image segmentation. In our study, we utilized the smallest image encoder (ViT-base) and did not fine-tune the image encoder to reduce computational burden. By employing larger backbone models and fine-tuning the image encoder, the model's capacity can be further enhanced to learn more features from underwater images. We plan to expand our training set to this scale in the future to further enhance AquaSAM's performance.

xx10    Muduo Xu, Jianhao Su, and Yutao Liu

## 5    Conclusion

In a nutshell, the success of SAM in natural image segmentation demonstrates the feasibility of building segmentation foundation models. This work makes the first attempt to adapt SAM to underwater image segmentation by fine-tuning the pre-trained model on underwater image datasets. We have achieved remarkable performance improvements across various tasks and image modalities. We hope that this work will inspire further studies to develop segmentation foundation models in the underwater image domain, and we anticipate significant advancements in the near future. Our code and trained model are publicly available, accompanied by a step-by-step tutorial on fine-tuning SAM on custom datasets.

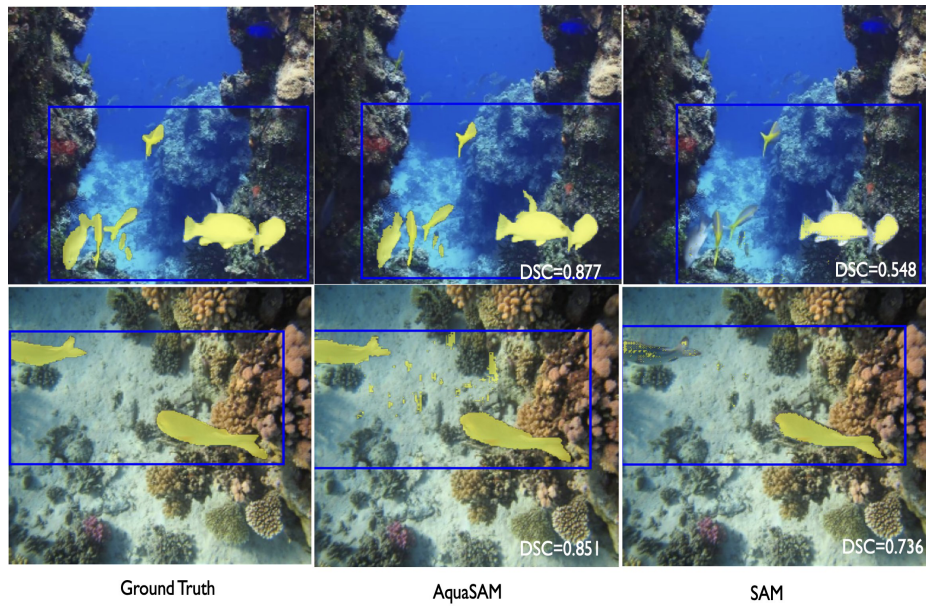

**Fig.5.** The Performance of AquaSAM and SAM when applied to multiple segmentation